\documentclass[runningheads]{llncs}
\usepackage{graphicx}
\usepackage{amsmath,amssymb,amsfonts}
\usepackage{lipsum}  
\usepackage{duckuments}
\usepackage{multirow}
\usepackage{adjustbox}
\usepackage{booktabs}
\usepackage{hyperref}
\usepackage[table,dvipsnames]{xcolor}
\usepackage{cleveref}
\usepackage[utf8]{inputenc}
\usepackage{soulutf8}
\usepackage{pifont}
\usepackage{amsmath}
\usepackage{amssymb}
\usepackage{color,soul}
\usepackage[dvipsnames]{xcolor}
\usepackage{gensymb}
\usepackage{mathtools}
\usepackage{orcidlink}
\usepackage{xcolor}
\hypersetup{
    colorlinks,
    linkcolor={red!80!black},
    citecolor={blue!50!black},
    urlcolor={blue!80!black}
}

\newcommand{\mytilde}{\raise.17ex\hbox{$\scriptstyle\mathtt{\sim}$}}

\DeclarePairedDelimiterX{\infdivx}[2]{[}{]}{#1\;\delimsize\|\;#2}

\def\ww{\mathbf{w}}
\def\xx{\mathbf{x}}

\def\WW{\mathbf{W}}

\def\lL{\mathcal{L}}

\def\wW{\mathcal{W}}
\def\xX{\mathcal{X}}

\newcommand{\result}[2]{\ensuremath{#1}\scriptsize{$\pm$\ensuremath{#2}}}

\begin{document}
\title{A Privacy-Preserving Walk in the Latent Space of Generative Models for Medical Applications}
\titlerunning{A Privacy-Preserving Walk in the Latent Space of Generative Models}
%
\author{Matteo Pennisi$^*$\inst{1}\orcidlink{0000-0002-6721-4383}
\and
Federica Proietto Salanitri\inst{1}\orcidlink{0000-0002-6122-4249}
\and
Giovanni Bellitto\inst{1}\orcidlink{0000-0002-1333-8348}
\and
Simone Palazzo\inst{1}\orcidlink{0000-0002-2441-0982}
\and
Ulas Bagci\inst{2}\orcidlink{0000-0001-7379-6829}
\and
Concetto Spampinato\inst{1}\orcidlink{0000-0001-6653-2577}}

\let\oldthefootnote=\thefootnote%
\def\thefootnote{*}\footnotetext{Corresponding Author}
\let\thefootnote=\oldthefootnote%

\authorrunning{M. Pennisi et al.}
%
\institute{PeRCeiVe Lab, University of Catania, Italy. \\ \url{http://www.perceivelab.com/}\and
Department of Radiology and BME, Northwestern University, Chicago, IL, USA}
\maketitle              
\begin{abstract}
Generative Adversarial Networks (GANs) have demonstrated their ability to generate synthetic samples that match a target distribution. However, from a privacy perspective, using GANs as a proxy for data sharing is not a safe solution, as they tend to embed near-duplicates of real samples in the latent space. Recent works, inspired by \emph{k-anonymity} principles, address this issue through sample aggregation in the latent space, with the drawback of reducing the dataset by a factor of $k$.
Our work aims to mitigate this problem by proposing a latent space navigation strategy able to generate diverse synthetic samples that may support effective training of deep models, while addressing privacy concerns in a principled way. Our approach leverages an \emph{auxiliary identity classifier} as a guide to non-linearly walk between points in the latent space, minimizing the risk of collision with near-duplicates of real samples. We empirically demonstrate that, given any random pair of points in the latent space, our walking strategy is safer than linear interpolation. 
We then test our path-finding strategy combined to \emph{k-same} methods and demonstrate, on two benchmarks for tuberculosis and diabetic retinopathy classification, that training a model using samples generated by our approach mitigate drops in performance, while keeping privacy preservation. Code is available at: \url{https://github.com/perceivelab/PLAN}

\end{abstract}
\section{Introduction}
The success of deep learning for medical data analysis has demonstrated its potential to become a core component of future diagnosis and treatment methodologies. 
However, in spite of the efforts devoted to improve data efficiency~\cite{kotia2021few}, the most effective models still rely on large datasets to achieve high accuracy and generalizability. 
An effective strategy for obtaining large and diverse datasets is to leverage collaborative efforts based on data sharing principles; 
however, current privacy regulations often hinder this possibility.
As a consequence, small private datasets are still used for training models that tend to overfit, introduce biases and generalize badly on other data sources addressing the same task~\cite{zech2018variable}.
As a mitigation measure, generative adversarial networks (GANs) have been proposed to synthesize highly-realistic images, extending existing datasets to include more (and more diverse) examples~\cite{Pennisi_2021_ICCV}, but they pose privacy concerns as real samples may be encoded in the latent space. 
\emph{K-same} techniques~\cite{jeon2022k,meden2018k} attempt to reduce this risk by following the \emph{k-anonymity} principle~\cite{sweeney2002k} and replacing real samples with synthetic aggregations of groups of $k$ samples. As a downside, these methods reduce the dataset size by a factor of $k$, which greatly limits their applicability.

To address this issue, we propose an approach, complementing \emph{k-same} techniques, for generating an extended variant of a dataset by sampling a privacy-preserving walk in the GAN latent space. Our method directly optimizes latent points, through the use of an \emph{auxiliary identity classifier}, which informs on the similarity between training samples and synthetic images corresponding to candidate latent points. This optimized navigation meets three key properties of data synthesis for medical applications: 1) \emph{equidistance}, encouraging the generation of diverse realistic samples suitable for model training;  2) \emph{privacy preservation}, limiting the possibility of recovering original samples, and, 3) \emph{class-consistency}, ensuring that synthesized samples contain meaningful clinical information.
To demonstrate the generalization capabilities of our approach, we experimentally evaluate its performance on two medical image tasks, namely, tuberculosis classification using the Shenzhen Hospital X-ray dataseet~\cite{candemir2013lung,jaeger2014two,jaeger2013automatic} and diabetic retinopathy classification on the APTOS dataset~\cite{aptos2019-blindness-detection}. On both tasks, our approach yields classification performance comparable to training with real samples and significantly better than existing \emph{k-same} techniques such as $k$-SALSA~\cite{jeon2022k}, while keeping the same robustness to membership inference attacks. 

\indent \textbf{Contributions}: 1) We present a latent space navigation approach that provides a large amount of diverse and meaningful images for model training; 2) We devise an optimization strategy of latent walks that enforces privacy; 3) We carry out several experiments on two medical tasks, demonstrating the effectiveness of our generative approach on model's training and its guarantees to privacy preservation.

\section{Related Work}
Conventional methods to protect identity in private images have involved modifying pixels through techniques like masking, blurring, and pixelation~\cite{7285017,pmid17295313}. However, these methods have been found to be insufficient for providing adequate privacy protection~\cite{refacing}.
As an alternative, GANs have been increasingly explored to synthesize high-quality images that preserve information from the original distribution, while disentangling and removing privacy-sensitive components~\cite{XuRZZQR19,yoon2018pategan}. However, these methods have been mainly devised for face images and cannot be directly applicable to medical images, since there is no clear distinction between identity and non-identity features~\cite{jeon2022k}.

Recent approaches, based on the \emph{k-same} framework~\cite{meden2018k}, employ GANs to synthesize clinically-valid medical images principle by aggregating groups of real samples into synthetic privacy-preserving examples~\cite{jeon2022k,pennisi2022gan}. In particular, k-SALSA~\cite{jeon2022k} uses GANs for generating retinal fundus images by proposing a local style alignment strategy to retain visual patterns of the original data. The main downside of these methods is that, in the strive to ensure privacy preservation following the \emph{k-anonymity}~\cite{sweeney2002k} principle, they significantly reduce the size of the original dataset.

Our latent navigation strategy complements these approaches by synthesizing large and diverse samples, suitable for downstream tasks. 
In general, latent space navigation in GANs manipulates the latent vectors to create new images with specific characteristics. While many works have explored this concept to control semantic attributes of generated samples~\cite{karras2020analyzing,brock2018large}, to the best of our knowledge, no method has tackled the problem from a privacy-preservation standpoint, especially on a critical domain such as medical image analysis.

\section{Method}
\begin{figure}[h!]
\centering
\includegraphics[width=\textwidth]{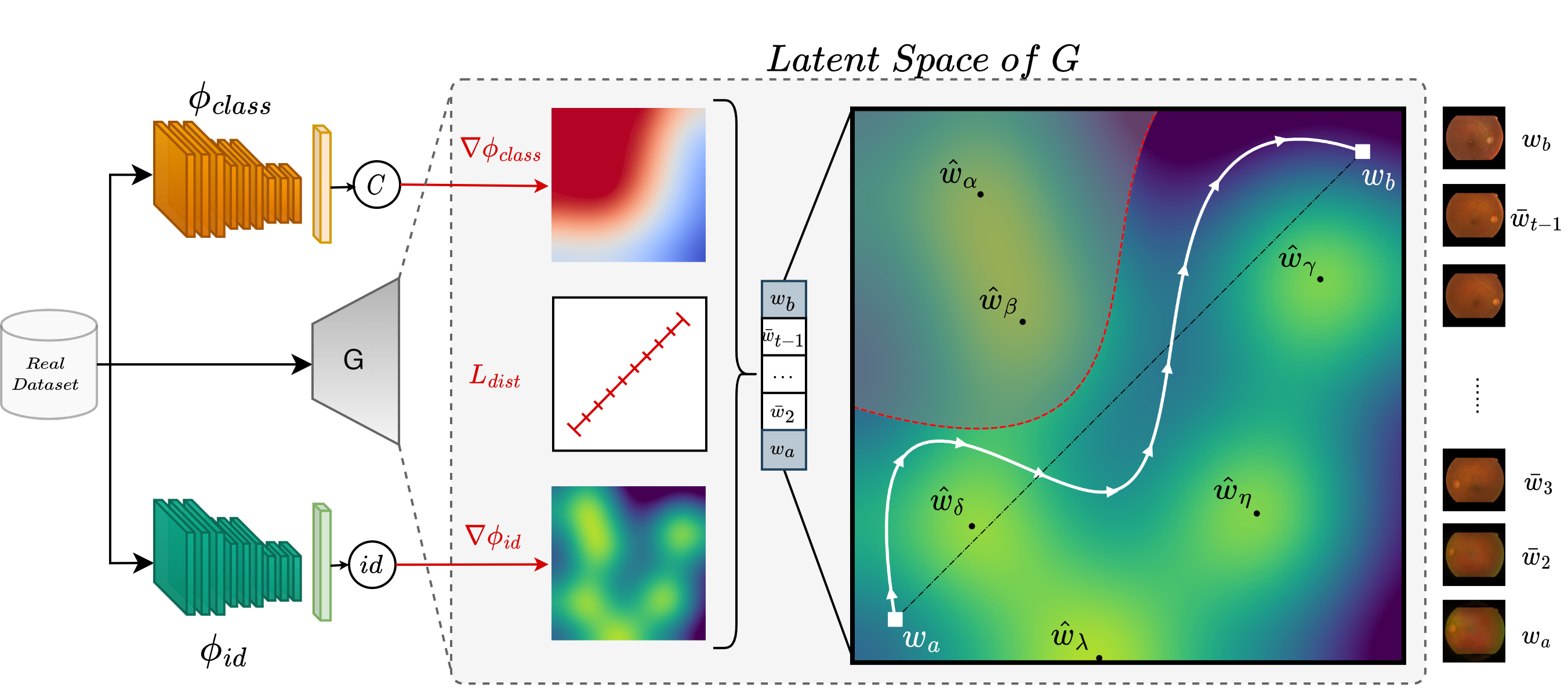}
\caption{\textbf{Overview of the PLAN approach}. Using real samples, we train a GAN, an \emph{identity classifier} $\phi_\text{id}$ and an \emph{auxiliary classifier} $\phi_\text{class}$. Given two arbitrary latent points, $\ww_a$ and $\ww_b$, PLAN employs $\phi_\text{id}$ and $\phi_\text{class}$ to gain information on latent space structure and generate a privacy-preserving navigation path (right image), from which synthetic samples can be sampled (far right images, zoom-in for details).}
\label{fig:architecture}
\end{figure}

The proposed \textbf{P}rivacy-preserving \textbf{LA}tent \textbf{N}avigation (\textbf{PLAN}) strategy envisages three separate stages: 1) GAN training using real samples; 2) latent privacy-preserving trajectory optimization in the GAN latent space; 3) privacy-preserving dataset synthesis for downstream applications. Fig.~\ref{fig:architecture} illustrates the overall framework and provides a conceptual interpretation of the optimization objectives. 

Formally, given a GAN generator $G: \wW \rightarrow \xX$, we aim to navigate its latent space $\wW$ to generate samples in image space $\xX$ in a privacy-preserving way, i.e., avoiding latent regions where real images might be embedded. The expected result is a synthetic dataset that is safe to share, while still including consistent clinical features to be used by downstream tasks (e.g., classification).

Our objective is to find a set of latent points $\bar{\mathcal{W}} \subset \wW$

from which it is safe to synthesize samples that are significantly different from training points: given the training set $\hat{\xX} \subset \xX$ and a metric $d$ on $\xX$, we want to find $\bar{\wW}$ such that $\min_{\xx \in \hat{\xX}} d\left( G\left(\bar{\ww}\right), \xx \right) > \delta$, $\forall \bar{\ww} \in \bar{\wW}$, for a sufficiently large $\delta$.
Manually searching for $\bar{\wW}$, however, may be unfeasible: generating a large $\bar{\wW}$ is computationally expensive, as it requires at least $|\bar{\wW}|$ forward passes through $G$, and each synthesized image should be compared to all training images;

moreover, randomly sampled latent points might not satisfy the above condition.

To account for latent structure, one could explicitly sample away from latent vectors corresponding to real data. Let $\hat{\wW_i} \subset \wW$ be the set of latent vectors that produce near-duplicates of a training sample $\xx_i \in \xX$, such that $G(\hat{\ww}_i) \approx \xx_i$, $\forall \hat{\ww}_i \in \hat{\wW_i}$.  
We can thus define $\hat{\wW} = \bigcup_{i=1}^N \hat{\wW}_i$ as the set of latent points corresponding to all $N$ samples of the training set: knowledge of $\hat{\wW}$ can be used to move the above constraint from $\xX$ to $\wW$, by finding $\bar{\wW}$ such that $\min_{\hat{\ww} \in \hat{\wW}} d\left( \bar{\ww}, \hat{\ww} \right) > \delta$, $\forall \bar{\ww} \in \bar{\wW}$. In practice, although $\hat{\wW_i}$ can be approximated through latent space projection~\cite{karras2020analyzing,alaluf2021restyle} from multiple initialization points, its cardinality $|\hat{\mathcal{W}_i}|$ cannot be determined \emph{a priori} as it is potentially unbounded.

From these limitations, we pose the search of seeking privacy-preserving latent points
as a trajectory optimization problem, constrained by a set of objectives that mitigate privacy risks and enforce sample variability and class consistency. Given two arbitrary latent points (e.g., provided by a \emph{k-same} aggregation method), $\ww_a,\ww_b \in \wW$, we aim at finding a latent trajectory $\bar{\WW}_T = \left[ \ww_a=\bar{\ww}_1, \bar{\ww}_2, \dots, \bar{\ww}_{T-1}, \ww_b=\bar{\ww}_T \right]$ that traverses the latent space from $\ww_a$ to $\ww_b$ in $T$ steps, such that none of its points can be mapped to any training sample. We design our navigation strategy to satisfy three requirements, which are then translated into optimization objectives:
\begin{enumerate}
    \item \textbf{Equidistance.} The distance between consecutive points in the latent trajectory should be approximately constant, to ensure sample diversity and mitigate mode collapse. We define the equidistance loss, $\lL_\text{dist}$, as follows:
\begin{equation}
\label{eq:path}
\mathcal{L}_\text{dist}= \sum_{i=1}^{T-1} \left\| \bar{\ww}_{i}, \bar{\ww}_{i+1} \right\|_2^2 
\end{equation}

where $\left\|\cdot\right\|_2$ is the $L_2$ norm. 
Note that without any additional constraint, $\lL_\text{dist}$ converges to the trivial solution of linear interpolation, which 

gives no guarantee that the path will not contain points belonging to $\hat{\mathcal{W}}$.

\item \textbf{Privacy preservation.} To navigate away from latent regions corresponding to real samples, we employ an auxiliary network $\phi_\text{id}$, trained on $\hat{\xX}$ to perform \emph{identity classification}. We then set the privacy preservation constraint by imposing that a sampled trajectory must maximize the uncertainty of $\phi_\text{id}$, thus avoiding samples that could be recognizable from the training set.
Assuming $\phi_\text{id}$ to be a neural network with as many outputs as the number of identities in the original dataset, this constraint can be mapped to a privacy-preserving loss, $\lL_\text{id}$, defined as the Kullback-Leibler divergence between the softmax probabilities of $\phi_{id}$ and the uniform distribution $\mathcal{U}$:
\begin{equation}
\label{eq:loss_id}
\lL_\text{id}=  \sum_{i=1}^{T} \text{KL} \infdivx{\phi_{id}(G(\bar{\ww}_i))}{\mathcal{U}(1/n_\text{id})}
\end{equation}
where $n_\text{id}$ is the number of identities.

This loss converges towards points with enhanced privacy, on which a trained classifier is maximally uncertain.

\item \textbf{Class consistency.} The latent navigation strategy, besides being privacy-preserving, needs to retain discriminative features to support training of downstream tasks on the synthetic dataset. 
In the case of a downstream classification task, given $\ww_a$ and $\ww_b$ belonging to the same class, all points along a trajectory between $\ww_a$ and $\ww_b$ should exhibit the visual features of that specific class. Moreover, optimizing the constraints in Eq.~\ref{eq:path} and Eq.~\ref{eq:loss_id} does not guarantee good visual quality, leading to privacy-preserving but useless synthetic samples. Thus, we add a third objective that enforces class-consistency on trajectory points. We employ an additional \emph{auxiliary classification network} $\phi_\text{class}$, trained to perform classification on the original dataset, to ensure that sampled latent points share the same visual properties (i.e., the same class) of $\ww_a$ and $\ww_b$. 
The corresponding loss $\lL_\text{class}$ is as follows: 
\begin{equation}
\label{eq:loss_label}
\lL_\text{class}=  \sum_{i=1}^{T} \text{CE}\left[ \phi_\text{class}(G(\bar{\ww}_i)), y \right]
\end{equation}
where $\text{CE}$ is the cross-entropy between the predicted label for each sample and the target class label $y$.
\end{enumerate}

\noindent Overall, the total loss for privacy-preserving latent navigation is obtained as:
\begin{equation}
\label{eq:loss_total}
\lL_\text{PLAN}= \mathcal{L}_{dist} + \lambda_1 \mathcal{L}_{id} + \lambda_2 \mathcal{L}_{label}
\end{equation}
where $\lambda_1$ and $\lambda_2$ weigh the three contributions.

In a practical application, we employ PLAN in conjunction with a privacy-preserving method that produces synthetic samples (e.g., a \emph{k-same} approach). We then navigate the latent space between random pairs of such samples, and increase the size of the dataset while retaining privacy preservation. The resulting extended set is then used to train a \emph{downstream classifier} $\phi_\text{down}$ on synthetic samples only. Overall, from an input set of $N$ samples, we apply PLAN to $N/2$ random pairs, thus sampling $TN/2$ new points.

\section{Experimental Results}
We demonstrate the effectiveness and privacy-preserving properties of our PLAN approach on two classification tasks, namely, tuberculosis classification and diabetic retinopathy (DR) classification. 

\subsection{Training and evaluation procedure}

\noindent\textbf{Data preparation.} For tuberculosis classification, we employ the Shenzhen Hospital X-ray set\footnote{This dataset was released by the National Library of Medicine, NIH, Bethesda, USA.}~\cite{candemir2013lung,jaeger2014two,jaeger2013automatic} that includes 662 frontal chest X-ray images (326 negatives and 336 positives).
For diabetic retinopathy classification, we use the APTOS fundus image dataset~\cite{aptos2019-blindness-detection} of retina images labeled by ophthalmologists with five grades of severity.
We downsample it by randomly selecting 950 images, equally distributed among classes, to simulate a typical scenario with low data availability (as in medical applications), where GAN-based synthetic sampling, as a form of augmentation, is more needed. 
All images are resized to 256$\times$256  and split into train, validation and test set with 70\%, 10\%, 20\% proportions.\\
\noindent\textbf{Baseline methods.} We evaluate our approach from a privacy-preserving perspective and by its capability to support downstream classification tasks. For the former, given the lack of existing methods for privacy-preserving GAN latent navigation, we compare PLAN to standard linear interpolation. After assessing privacy-preserving performance, we measure the impact of our PLAN sampling strategy when combined to k-SALSA~\cite{jeon2022k} and the latent cluster interpolation approach from~\cite{pennisi2022gan} (LCI in the following) on the two considered tasks. \\
\noindent\textbf{Implementation details.} We employ StyleGAN2-ADA~\cite{karras2020training} as GAN model for all baselines, trained in a label-conditioned setting on the original training sets.
For all classifiers ($\phi_\text{id}$, $\phi_\text{class}$ and $\phi_\text{down}$) we employ a ResNet-18 network~\cite{he2016deep}. Classifiers $\phi_\text{id}$ and $\phi_\text{class}$ are trained on the original training set, while $\phi_\text{down}$ (i.e., the task classifier, one for each task) is trained on synthetic samples only. For $\phi_\text{id}$, we apply standard data augmentation (e.g., horizontal flip, rotation) and add five GAN projections for each identity, to mitigate the domain shift between real and synthetic images. $\phi_\text{down}$ is trained with a learning rate of 0.001, a batch size of 32, for 200 (Shenzhen) and 500 (APTOS) epochs. Model selection is carried out at the best validation accuracy, and results are averaged over 5 runs.
When applying PLAN on a pair of latent points, we initialize a trajectory of $T=50$ points through linear interpolation, and optimize Eq.~\ref{eq:loss_total} for 100 steps using Adam with a learning rate of 0.1; $\lambda_1$ and $\lambda_2$ are set to 0.1 and 1, respectively. Experiments are performed on an NVIDIA RTX 3090.

\subsection{Results}
To measure the privacy-preserving properties of our approach, we employ the \emph{membership inference attack} (MIA)~\cite{shokri2017membership}, which attempts to predict if a sample was used in a classifier's training set. We use  attacker model and settings defined in~\cite{nasr2018machine,jia2019memguard}, training the attacker on 30\% of the training set (seen by PLAN through $\phi_\text{id}$ and $\phi_\text{class}$) and 30\% of the test set (unseen by PLAN); as a test set for MIA, we reserve 60\% of the original test set, leaving 10\% as a validation set to select the best attacker. Ideally, if the model preserves privacy, the attacker achieves chance performance (50\%), showing inability to identify samples used for training. We also report the FID of the generated dataset, to measure its level of realism, and the mean of the minimum LPIPS~\cite{zhang2018unreasonable} (``$mmL$'' for short) distances between each generated sample and its closest real image, to measure how generated samples differ from real ones. We compare PLAN to a linear interpolation between arbitrary pairs of start and end latent points, and compute the above measures on the images corresponding to the latent trajectories obtained by two approaches. We also report the results of the classifier trained on real data to provide additional bounds for both classification accuracy and privacy-preserving performance. 
\begin{table}[ht!]
    \centering
    \rowcolors{2}{gray!10}{white}
    \renewcommand{\arraystretch}{1.2}
    \caption{Comparison between the downstream classifier ($\phi_\text{down}$) model trained with real samples and those trained with synthetic samples generated from the linear path and privacy path, respectively.}
    \label{tab:method}
    \resizebox{1.\linewidth}{!}{%
    \setlength{\tabcolsep}{2pt}
    \begin{tabular}{l|cccc|cccc}
        \toprule
                & \multicolumn{4}{c}{\textbf{Shenzhen}}                                                 & \multicolumn{4}{c}{\textbf{Aptos}}\\
        \midrule
                &  Acc. (\%){\scriptsize ($\uparrow$)}                & MIA {\scriptsize ($\downarrow$)}                 & FID {\scriptsize ($\downarrow$)}        & mmL {\scriptsize ($\uparrow$)}  & Acc. (\%){\scriptsize ($\uparrow$)}                & MIA {\scriptsize ($\downarrow$)}                 & FID {\scriptsize ($\downarrow$)}        & mmL {\scriptsize ($\uparrow$)} \\
        \midrule
        Real    & \result{81.23}{1.03}      & \result{71.41}{3.59}      & --        & --                               &\result{50.74}{2.85}     & \result{73.30}{4.04}      & --           & -- \\
        Linear  & \result{82.14}{1.40}      & \result{56.28}{1.60}      & 63.85     & 0.125                            &\result{41.58}{2.11}     & \result{50.53}{3.06}      & \textbf{85.17}        & 0.118\\
        PLAN & \result{\textbf{83.85}}{1.33}      & \result{\textbf{50.13}}{3.99}      & \textbf{63.22}     & \textbf{0.159}                            &\result{\textbf{46.95}}{3.06}     & \result{\textbf{48.51}}{2.85}      & 90.81        & \textbf{0.131}\\
        \bottomrule
    \end{tabular}
    }
\end{table}

Results in Table~\ref{tab:method} demonstrate that our approach performs similarly to training with real data, but with higher accuracy with respect to the linear baseline. Privacy-preserving results, measured through MIA and $mmL$, demonstrate the reliability of our PLAN strategy in removing sensitive information, reaching the ideal lower bound of MIA accuracy. \\
\begin{figure}[h!]
\centering
\includegraphics[width=1.\textwidth]{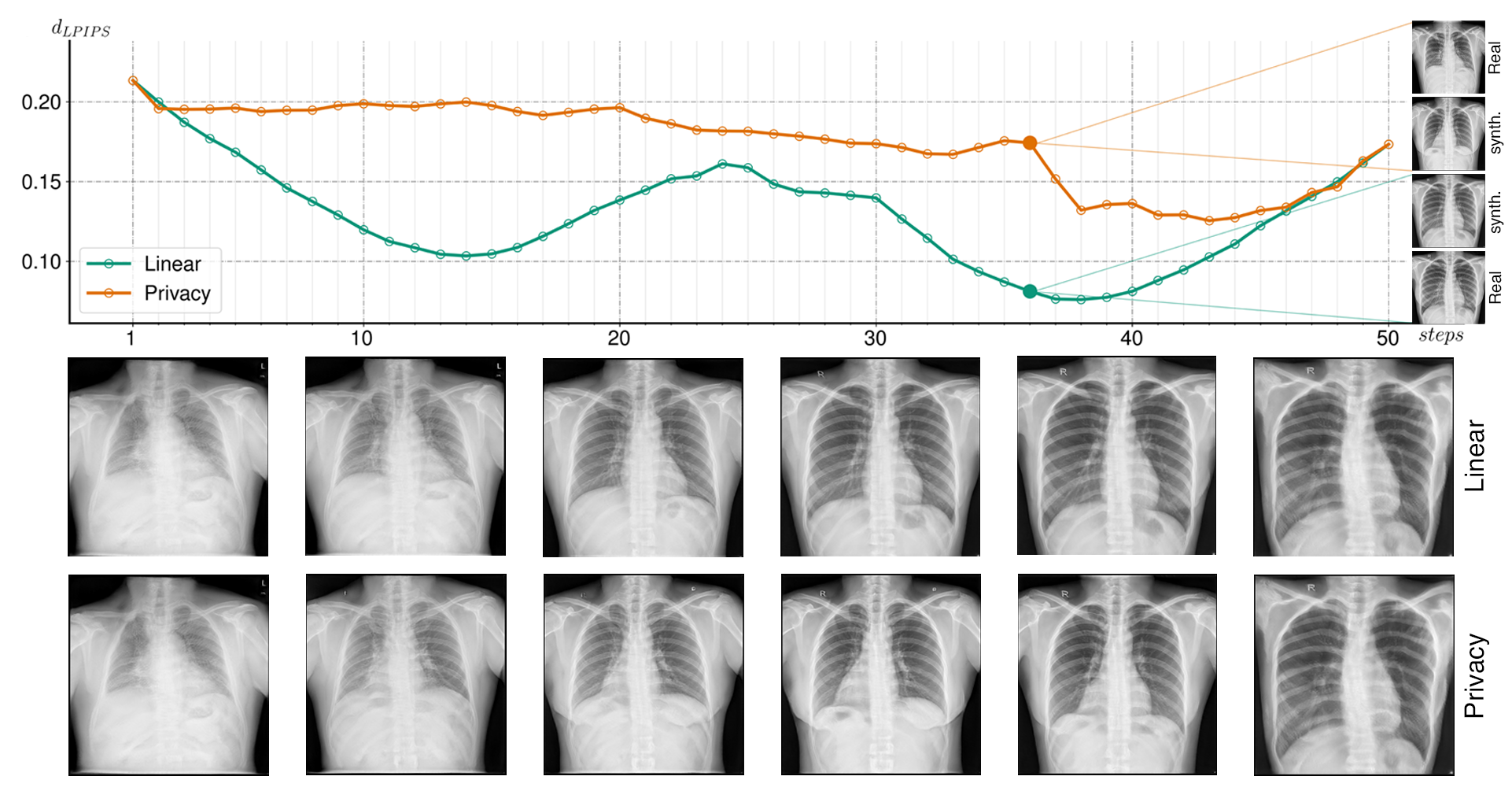}
\caption{\textbf{Linear vs PLAN navigation between two arbitrary points.} For each step of the latent trajectory, we compute the LPIPS distance between each synthetic sample and its closest real image. On the right, a qualitative comparison of images at step 35 and their closest real samples: the synthetic image obtained with PLAN differs significantly from its closest real sample; in linear interpolation, synthetic and real samples look similar. Bottom images show synthetic samples generated by linear interpolation and PLAN at the same steps (zoom-in for details).}
\label{fig:path images}
\end{figure}
Fig.~\ref{fig:path images} shows how, for given start and end points, PLAN-generated samples keep high quality but differ significantly from real samples, while latent linear interpolation may lead to near-duplicates. This is confirmed by the higher LPIPS distance between generated samples and the most similar real samples for PLAN.\\
After verifying the generative and privacy-preserving capabilities of our approach, we evaluate its contribution to classification accuracy when combined with existing \emph{k-same} methods, namely k-SALSA~\cite{jeon2022k} and LCI~\cite{pennisi2022gan}. 
Both methods apply latent clustering to synthesize a privacy-preserving dataset, but exhibit low  performance transferability to classification tasks, due to the reduced  size of the resulting synthetic dataset.
We carry out these experiments on APTOS, using $k=5$ and $k=10$, for comparison with~\cite{jeon2022k}\footnote{Values of $k$ smaller than 5 led to vulnerabilities to MIA on APTOS, as shown in \cite{jeon2022k}.}.
Results are given in Table~\ref{tab:sota} and show how our PLAN strategy enhances performance of the two baseline methods, reaching performance similar to training the retinopathy classifier with real samples (i.e., 50.74 on real data vs 44.95 when LCI~\cite{pennisi2022gan} is combined with PLAN) and much higher than the variants without PLAN.  We also measured MIA accuracy between the variants with and without PLAN, and we did not observe significant change among the different configurations: accuracy was at the chance level in all cases, suggesting their privacy-preserving capability.

\begin{table}[ht!]
    \centering
    \rowcolors{3}{white}{gray!10}
    \setlength{\tabcolsep}{2pt}
    \renewcommand{\arraystretch}{1.5}
    \caption{Impact of our navigation strategy on k-same methods on the APTOS dataset. Performance are reported in terms of accuracy.}
    \begin{tabular}{l|cc|cc}
    \toprule

              &$k$-SALSA~\cite{jeon2022k} & \textbf{$k$-SALSA}                        &LCI~\cite{pennisi2022gan} & \textbf{LCI} \\
              & & \textbf{+PLAN} & & \textbf{+PLAN}\\
    \bottomrule
         k\;=\;5  & \result{25.58}{6.32}                     & \result{\textbf{36.59}}{3.48}  & \result{38.74}{4.51}                      & \result{\textbf{43.16}}{2.71}\\
         k\;=\;10 & \result{27.47}{3.42}                     & \result{\textbf{34.21}}{1.62}  & \result{36.42}{3.77}                      & \result{\textbf{44.95}}{1.61}\\
    \bottomrule
    \end{tabular}
    \label{tab:sota}
\end{table}

\section{Conclusion}
We presented PLAN, a latent space navigation strategy designed to reduce privacy risks when using GANs for training models on synthetic data. 
Experimental results, on two medical image analysis tasks, demonstrate how PLAN is robust to membership inference attacks while effectively supporting model training with performance comparable to training on real data. 
Furthermore, when PLAN is combined with state-of-the-art \emph{k-anonymity} methods, we observe a mitigation of performance drop while maintaining privacy-preservation properties.
Future research directions will address the scalability of the method to large datasets with a high number of identities, as well as  learning latent trajectories with arbitrary length to maximize privacy-preserving and augmentation properties of the synthetic datasets.

\subsubsection{Acknowledgements} This research was supported by PNRR MUR project PE0000013-FAIR. Matteo Pennisi is a PhD student enrolled in the National PhD in Artificial Intelligence, XXXVII cycle, course on Health and life sciences, organized by Università Campus Bio-Medico di Roma.
%
%
%
\bibliographystyle{splncs04}
\bibliography{biblio}

\begin{thebibliography}{10}
\providecommand{\url}[1]{\texttt{#1}}
\providecommand{\urlprefix}{URL }
\providecommand{\doi}[1]{https://doi.org/#1}

\bibitem{refacing}
Abramian, D., Eklund, A.: Refacing: Reconstructing anonymized facial features
  using {GANS}. In: 16th {IEEE} International Symposium on Biomedical Imaging,
  {ISBI} 2019, Venice, Italy, April 8-11, 2019. pp. 1104--1108. {IEEE} (2019)

\bibitem{alaluf2021restyle}
Alaluf, Y., Patashnik, O., Cohen-Or, D.: Restyle: A residual-based stylegan
  encoder via iterative refinement. In: Proceedings of the IEEE/CVF
  International Conference on Computer Vision. pp. 6711--6720 (2021)

\bibitem{pmid17295313}
Bischoff-Grethe, A., Ozyurt, I.B., Busa, E., Quinn, B.T., Fennema-Notestine,
  C., Clark, C.P., Morris, S., Bondi, M.W., Jernigan, T.L., Dale, A.M., Brown,
  G.G., Fischl, B.: {{A} technique for the deidentification of structural brain
  {M}{R} images}. Hum Brain Mapp  \textbf{28}(9),  892--903 (Sep 2007)

\bibitem{brock2018large}
Brock, A., Donahue, J., Simonyan, K.: Large scale gan training for high
  fidelity natural image synthesis. arXiv preprint arXiv:1809.11096  (2018)

\bibitem{candemir2013lung}
Candemir, S., Jaeger, S., Palaniappan, K., Musco, J.P., Singh, R.K., Xue, Z.,
  Karargyris, A., Antani, S., Thoma, G., McDonald, C.J.: Lung segmentation in
  chest radiographs using anatomical atlases with nonrigid registration. IEEE
  transactions on medical imaging  \textbf{33}(2),  577--590 (2013)

\bibitem{he2016deep}
He, K., Zhang, X., Ren, S., Sun, J.: {Deep residual learning for image
  recognition}. In: Proceedings of the IEEE conference on Computer Vision and
  Pattern Recognition (2016)

\bibitem{jaeger2014two}
Jaeger, S., Candemir, S., Antani, S., W{\'a}ng, Y.X.J., Lu, P.X., Thoma, G.:
  Two public chest x-ray datasets for computer-aided screening of pulmonary
  diseases. Quantitative imaging in medicine and surgery  \textbf{4}(6), ~475
  (2014)

\bibitem{jaeger2013automatic}
Jaeger, S., Karargyris, A., Candemir, S., Folio, L., Siegelman, J., Callaghan,
  F., Xue, Z., Palaniappan, K., Singh, R.K., Antani, S., et~al.: Automatic
  tuberculosis screening using chest radiographs. IEEE transactions on medical
  imaging  \textbf{33}(2),  233--245 (2013)

\bibitem{jeon2022k}
Jeon, M., Park, H., Kim, H.J., Morley, M., Cho, H.: k-salsa: k-anonymous
  synthetic averaging of retinal images via local style alignment. In: Computer
  Vision--ECCV 2022: 17th European Conference, Tel Aviv, Israel, October
  23--27, 2022, Proceedings, Part XXI. pp. 661--678. Springer (2022)

\bibitem{jia2019memguard}
Jia, J., Salem, A., Backes, M., Zhang, Y., Gong, N.Z.: Memguard: Defending
  against black-box membership inference attacks via adversarial examples. In:
  Proceedings of the 2019 ACM SIGSAC conference on computer and communications
  security. pp. 259--274 (2019)

\bibitem{karras2020training}
Karras, T., Aittala, M., Hellsten, J., Laine, S., Lehtinen, J., Aila, T.:
  Training generative adversarial networks with limited data. Advances in
  Neural Information Processing Systems  \textbf{33},  12104--12114 (2020)

\bibitem{karras2020analyzing}
Karras, T., Laine, S., Aittala, M., Hellsten, J., Lehtinen, J., Aila, T.:
  Analyzing and improving the image quality of stylegan. In: Proceedings of the
  IEEE/CVF conference on computer vision and pattern recognition. pp.
  8110--8119 (2020)

\bibitem{aptos2019-blindness-detection}
Karthik, Maggie, S.D.: Aptos 2019 blindness detection (2019),
  \url{https://kaggle.com/competitions/aptos2019-blindness-detection}

\bibitem{kotia2021few}
Kotia, J., Kotwal, A., Bharti, R., Mangrulkar, R.: Few shot learning for
  medical imaging. Machine learning algorithms for industrial applications pp.
  107--132 (2021)

\bibitem{meden2018k}
Meden, B., Emer{\v{s}}i{\v{c}}, {\v{Z}}., {\v{S}}truc, V., Peer, P.:
  k-same-net: k-anonymity with generative deep neural networks for face
  deidentification. Entropy  \textbf{20}(1), ~60 (2018)

\bibitem{nasr2018machine}
Nasr, M., Shokri, R., Houmansadr, A.: Machine learning with membership privacy
  using adversarial regularization. In: Proceedings of the 2018 ACM SIGSAC
  conference on computer and communications security. pp. 634--646 (2018)

\bibitem{Pennisi_2021_ICCV}
Pennisi, M., Palazzo, S., Spampinato, C.: Self-improving classification
  performance through gan distillation. In: Proceedings of the IEEE/CVF
  International Conference on Computer Vision (ICCV) Workshops. pp. 1640--1648
  (October 2021)

\bibitem{pennisi2022gan}
Pennisi, M., Proietto~Salanitri, F., Palazzo, S., Pino, C., Rundo, F.,
  Giordano, D., Spampinato, C.: Gan latent space manipulation and aggregation
  for federated learning in medical imaging. In: Distributed, Collaborative,
  and Federated Learning, and Affordable AI and Healthcare for Resource Diverse
  Global Health: Third MICCAI Workshop, DeCaF 2022, and Second MICCAI Workshop,
  FAIR 2022, Held in Conjunction with MICCAI 2022, Singapore, September 18 and
  22, 2022, Proceedings. pp. 68--78. Springer (2022)

\bibitem{7285017}
Ribaric, S., Pavesic, N.: An overview of face de-identification in still images
  and videos. In: 2015 11th IEEE International Conference and Workshops on
  Automatic Face and Gesture Recognition (FG). vol.~04, pp.~1--6 (2015)

\bibitem{shokri2017membership}
Shokri, R., Stronati, M., Song, C., Shmatikov, V.: Membership inference attacks
  against machine learning models. In: 2017 IEEE symposium on security and
  privacy (SP). pp. 3--18. IEEE (2017)

\bibitem{sweeney2002k}
Sweeney, L.: k-anonymity: A model for protecting privacy. International journal
  of uncertainty, fuzziness and knowledge-based systems  \textbf{10}(05),
  557--570 (2002)

\bibitem{XuRZZQR19}
Xu, C., Ren, J., Zhang, D., Zhang, Y., Qin, Z., Ren, K.: Ganobfuscator:
  Mitigating information leakage under gan via differential privacy. IEEE
  Trans. Inf. Forensics Secur.  \textbf{14}(9),  2358--2371 (2019)

\bibitem{yoon2018pategan}
Yoon, J., Jordon, J., van~der Schaar, M.: {PATE}-{GAN}: Generating synthetic
  data with differential privacy guarantees. In: International Conference on
  Learning Representations (2019)

\bibitem{zech2018variable}
Zech, J.R., Badgeley, M.A., Liu, M., Costa, A.B., Titano, J.J., Oermann, E.K.:
  Variable generalization performance of a deep learning model to detect
  pneumonia in chest radiographs: a cross-sectional study. PLoS medicine
  \textbf{15}(11),  e1002683 (2018)

\bibitem{zhang2018unreasonable}
Zhang, R., Isola, P., Efros, A.A., Shechtman, E., Wang, O.: The unreasonable
  effectiveness of deep features as a perceptual metric. In: Proceedings of the
  IEEE conference on computer vision and pattern recognition. pp. 586--595
  (2018)

\end{thebibliography}

\end{document}